\documentclass{article} 
\usepackage[table]{xcolor}
\usepackage{colm2024_conference}
\usepackage{microtype}
\usepackage{hyperref}
\usepackage{url}
\usepackage{booktabs}
\usepackage{amsmath}
\usepackage{multirow}
\usepackage{multicol}
\usepackage{CJKutf8}
\usepackage{tcolorbox}
\usepackage{tabularx}
\usepackage{cleveref}
\usepackage{subcaption}

\definecolor{darkblue}{rgb}{0, 0, 0.5}
\hypersetup{colorlinks=true, citecolor=darkblue, linkcolor=darkblue, urlcolor=darkblue}
\newcommand{\ignore}[1]{}

\title{DianJin-OCR-R1: Enhancing OCR Capabilities via a Reasoning-and-Tool Interleaved Vision-Language Model}

\author{Qian Chen, Xianyin Zhang, Lifan Guo, Feng Chen, Chi Zhang \\
Qwen DianJin Team, Alibaba Cloud Computing
}

\colmfinalcopy 
\begin{document}

\maketitle

\begin{CJK*}{UTF8}{gkai}

\begin{abstract}
Recent advances in vision-language models (VLMs) have enabled end-to-end document parsing and understanding, achieving strong performance on diverse optical character recognition (OCR) tasks. However, VLMs are prone to generate words that do not exist in the input image due to over-reliance on language priors. By contrast, traditional OCR models, whose architectures are tailored for specific recognition tasks, often achieve stronger fine-grained visual perception with fewer hallucinations, but they typically lack the contextual semantic understanding and reasoning capabilities needed in more challenging cases. To bridge this gap, we propose DianJin-OCR-R1, a reasoning-enhanced framework for recognition that trains VLMs in a reasoning-and-tool interleaved paradigm. Our DianJin-OCR-R1 model first recognizes the content in the input image through its own OCR capabilities, and then calls other expert models for extra results as references. After that, it is guided to "look again" at the image and compare its own recognized content with other results to find errors or omissions. Finally, it integrates all available evidence to generate a more accurate output. This design empowers the model to learn how to implicitly re-focus on the visual input and effectively leverage the results of other expert models for better performance. We evaluate our DianJin-OCR-R1 model on ReST and OmniDocBench, where it consistently outperforms both its non-reasoning counterparts and expert models, demonstrating the effectiveness of our method.
\end{abstract}

\section{Introduction}
\label{sec:introduction}
Extracting structured content from document images containing intertwined elements such as plain texts, tables, and formulas is crucial for advancing AI-assisted document understanding, and it is also one of the most critical steps in numerous downstream applications such as Retrieval-Augmented Generation (RAG). Traditional OCR systems like PP-Structure \cite{li2022ppstructure, xcui2025paddleocr} train expert models for different tasks separately, such as SLANet \cite{li2022ppstructure} for table recognition and PP-FormulaNet \cite{liu2025pp} for formula recognition. Compared to PP-Structure with complex procedures, vision-language models (VLMs) can process various document elements in an unified, elegant, and end-to-end paradigm. In particular, expert VLMs such as GOT \cite{wei2024general} are trained on various OCR tasks to directly generate structured content from document images.

Despite these advancements, these models still exhibit their respective limitations. VLMs are prone to generate words that do not exist in the input image. One of the possible cause of this phenomenon is that they have learned language patterns from training datasets used in the language models, instead of carefully recognizing the actual content in the input images. In contrast, expert OCR models whose architectures are tailored for specific OCR tasks, rarely have hallucination issues, and typically only recognize several words incorrectly in challenging cases. However, they rely solely on visual information. When several words in an image are difficult to recognize, they cannot leverage contextual semantic information to obtain better answers. 

\begin{CJK*}{UTF8}{gkai}
\begin{table}[t]
    \caption{A challenging example in the seal recognition task. Red text indicates the errors. VLMs (e.g., Qwen2.5-VL-7B-Instruct \cite{bai2025qwen2}) are prone to generate words that are not grounded in the input image, which can be attributed to the fact that they have learned language patterns from training datasets. In contrast, expert OCR models (e.g., PP-StructureV3 \cite{xcui2025paddleocr}) rely primarily on visual features and lack contextual semantic understanding, which can lead to multiple word-level recognition errors in such challenging case.}
    \label{tab:example}
    \centering
        \begin{tabularx}{\textwidth}{l|X|X}
            \toprule
            \multirow{4}{*}{\includegraphics[width=3.5cm]{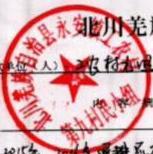}} & Prompt & 请识别图中印章的内容，直接输出内容即可。 \\
            \cmidrule{2-3}
            \cmidrule{2-3}
            & Qwen2.5-VL-7B-Instruct & 北川羌族自治县\textcolor{red}{农村信用合作联社永安分社} \\
            \cmidrule{2-3}
            & PP-StructureV3 & 北川\textcolor{red}{芜}族自治县永安\textcolor{red}{断}工农\textcolor{red}{行} 第九村民小组 \\
            \cmidrule{2-3}
            & Ground-truth & 北川羌族自治县永安镇工农村 第九村民小组 \\
            \bottomrule
        \end{tabularx}
\end{table}
\end{CJK*}

To synergize the advantages of both approaches while addressing their limitations, we present Dianjin-OCR-R1, a novel reasoning-enhanced framework based on VLMs, following a reasoning-and-tool interleaved paradigm. Inspire by reasoning models \cite{guo2025deepseek} that are trained to articulate their problem-solving process through a Chain-of-Thought (CoT) manner, our Dianjin-OCR-R1 model first recognizes the content in the input image through its own OCR capabilities, and then calls other expert models for extra results as references. After that, it is guided to "look again" at the image implicitly and compare its own recognized content with other results to find errors or omissions. Finally, it integrates all available evidence to generate a more accurate output. Our method guides VLMs to learn how to implicitly re-focus on the visual input during the reasoning process, thus enhancing their attention to visual input. Meanwhile, the additional results from other expert models provide references that enable the VLM to better self-validate its predictions, thereby alleviating—at least in part—the tendency to generate words that is not present in the image. In addition, compared with post-training the VLMs, expert models are easier to iterate which allows VLMs to improve their performance at a lower cost. 

We train our DianJin-OCR-R1 models based on Qwen2.5-VL-7B-Instruct \cite{bai2025qwen2} by Supervised Fine-Tuning (SFT) and Reinforcement Fine-Tuning (RFT). We apply Group Relative Policy Optimization (GRPO) \cite{shao2024deepseekmath}, a reinforcement learning algorithm that introduces two reward signals: a format reward to encourage structured outputs and an accuracy reward to promote answer correctness. These mechanisms guide the model to produce coherent, verifiable reasoning chains and reliable answers. 

Our main contributions can be summarized as follows:
\begin{itemize}
    \item We propose a novel reasoning-and-tool interleaved framework, Dianjin-OCR-R1, balancing the understanding and reasoning ability of VLMs and the recognition ability of expert models. 
    \item We conduct a thorough experiment on benchmarks including ReST\cite{yu2023icdar} and OmniDocBench\cite{ouyang2025omnidocbench}. Experimental results show that our Dianjin-OCR-R1 models consistently outperform their non-reasoning counterparts and expert models, which proves the effectiveness of our method.
    \item Our DianJin-OCR-R1 models only require the replacement or iteration of expert models to improve their performance, without requiring other post-training of VLMs, which allows us to better utilize resources.
    \item Our method can be used to generate more and better synthetic data through the most advanced model on the market, which will further advance the development of document parsing.
\end{itemize}

\section{Related Work}
\subsection{General VLMs}
General VLMs such as GPT series \cite{yang2023dawn}, Gemini series \cite{team2024gemini}, Qwen-VL series \cite{Qwen-VL, wang2024qwen2, bai2025qwen2}, MiniCPM series \cite{yao2024minicpm}, Intern-VL series \cite{chen2024internvl, chen2024expanding, zhu2025internvl3}, and Seed-VL series \cite{guo2025seed1} have demonstrated encouraging results in document parsing and understanding, and do not require training for specific tasks. However, general VLMs focus on visual reasoning performance, e.g., Visual Question Answering (VQA), and their capabilities in perception are not as strong.

\subsection{Expert VLMs}
Expert VLMs such as GOT \cite{wei2024general}, MonkeyOCR \cite{liu2024textmonkey, li2025monkeyocr}, mPLUG-DocOwl \cite{ye2307mplug, ye2023ureader}, olmOCR \cite{poznanski2025olmocr}, Ocean-OCR \cite{chen2025ocean}, and Mistral-OCR\footnote{\url{https://mistral.ai/fr/news/mistral-ocr}} are specifically designed and trained for document parsing. They used an innovative unified model that processes various document elements. However, they still suffer from hallucinations.

\subsection{Expert OCR Models}
Researchers usually trained different expert OCR models for different OCR tasks separately. PP-OCR series \cite{du2020pp, du2021pp, li2022pp, liu2025pp} focus on multi-scenario and multi-lingual text detection and recognition tasks. TableMaster \cite{cao2025tablemaster}, SLANet \cite{li2022ppstructure}, and UniTable \cite{peng2024unitable} are proposed to recognize the structure and content of tables. LaTeX-OCR\footnote{\url{https://github.com/lukas-blecher/LaTeX-OCR}}, PP-FormulaNet \cite{liu2025pp}, and UniMERNet \cite{wang2024unimernet} are designed for formula recognition task. However, one inconvenient aspect is that researchers often have to switch between different models to meet various recognition requirements.

\subsection{Reasoning Models}
Recent advances in reasoning models, optimized through reinforcement learning, highlight the potential for foundation models to enhance inference-time scaling and achieve higher levels of intelligence, e.g., OpenAI o3\footnote{\url{https://openai.com/index/introducing-o3-and-o4-mini}}, DeepSeek-R1 \cite{guo2025deepseek}. Following the success of text-only reasoning models, researches \cite{shen2025vlm, jiang2025vlm} in VLM reasoning have emerged, focusing on both effective multimodal CoT structures and high-quality training data construction methods. However, the majority of them are developed for VQA tasks, with few intended for recognition tasks.

\section{method}
\label{sec:method}

\begin{figure}[t]
    \centering
    \includegraphics[width=1.0\columnwidth]{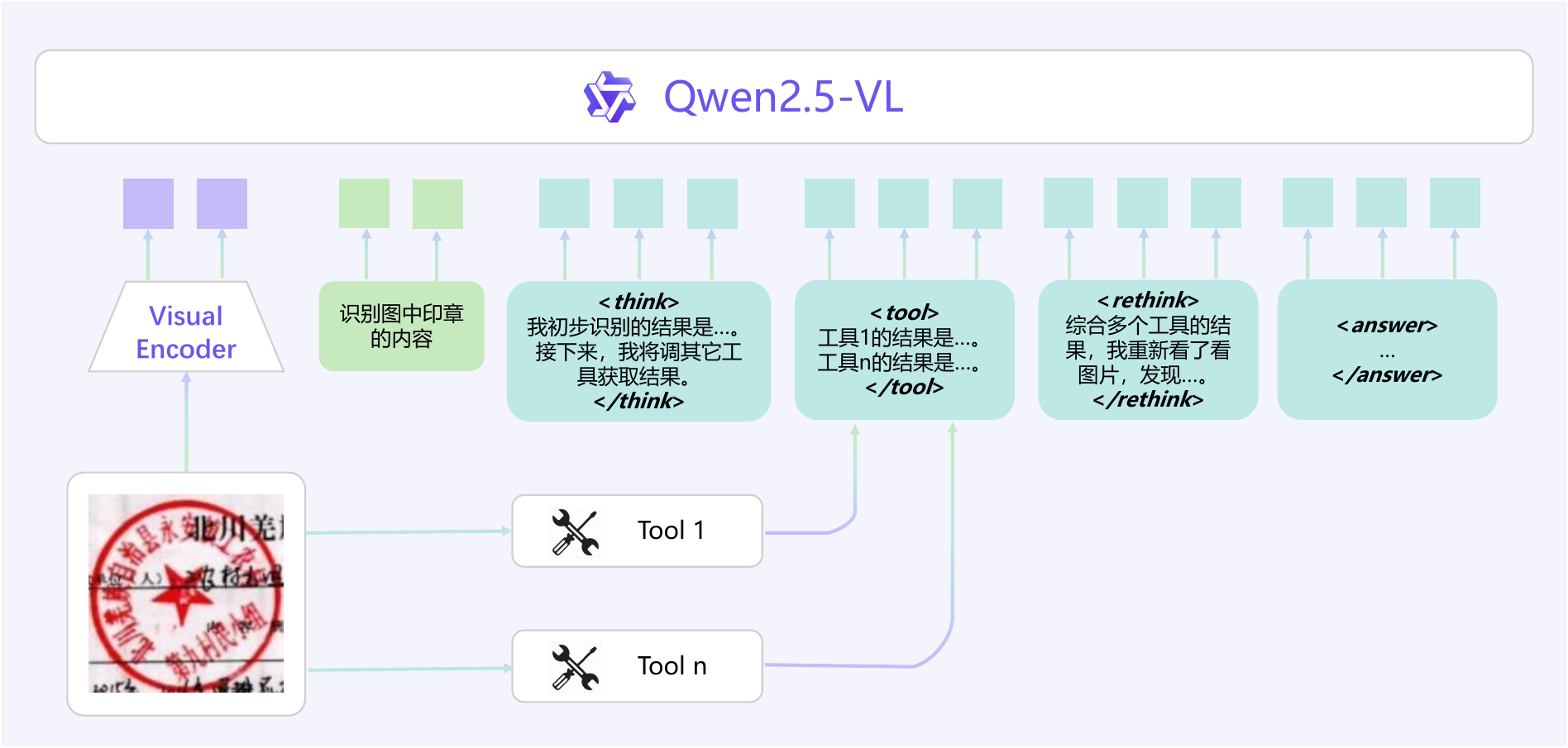}
    \caption{The visualization of the reason pipeline of our proposed method.}
    \label{fig:model}
\end{figure}

\subsection{Reasoning Paradigm}
Our proposed DianJin-OCR-R1, focusing on recognition tasks, fosters a straightforward reasoning paradigm in VLMs, enabling a more transparent and verifiable reasoning process. As shown in \Cref{fig:model}, given an image and a recognition instruction, DianJin-OCR-R1 first enables the VLM to recognize the content in the input image using its own OCR capability (within <think> and </think> tags). Then, it calls other expert models or tools, using their results as references or supplementary information (within <tool> and </tool> tags). Next, it guide the VLM to "look again" at the image, comprehensively analyzes both its own results and those from other models, reflects on whether it made any mistakes or missed anything during recognition (within <rethink> and </rethink> tags). Finally, it integrates all available evidence to generate a more accurate output.

 \begin{figure}[t]
    \centering
    \includegraphics[width=1.0\columnwidth]{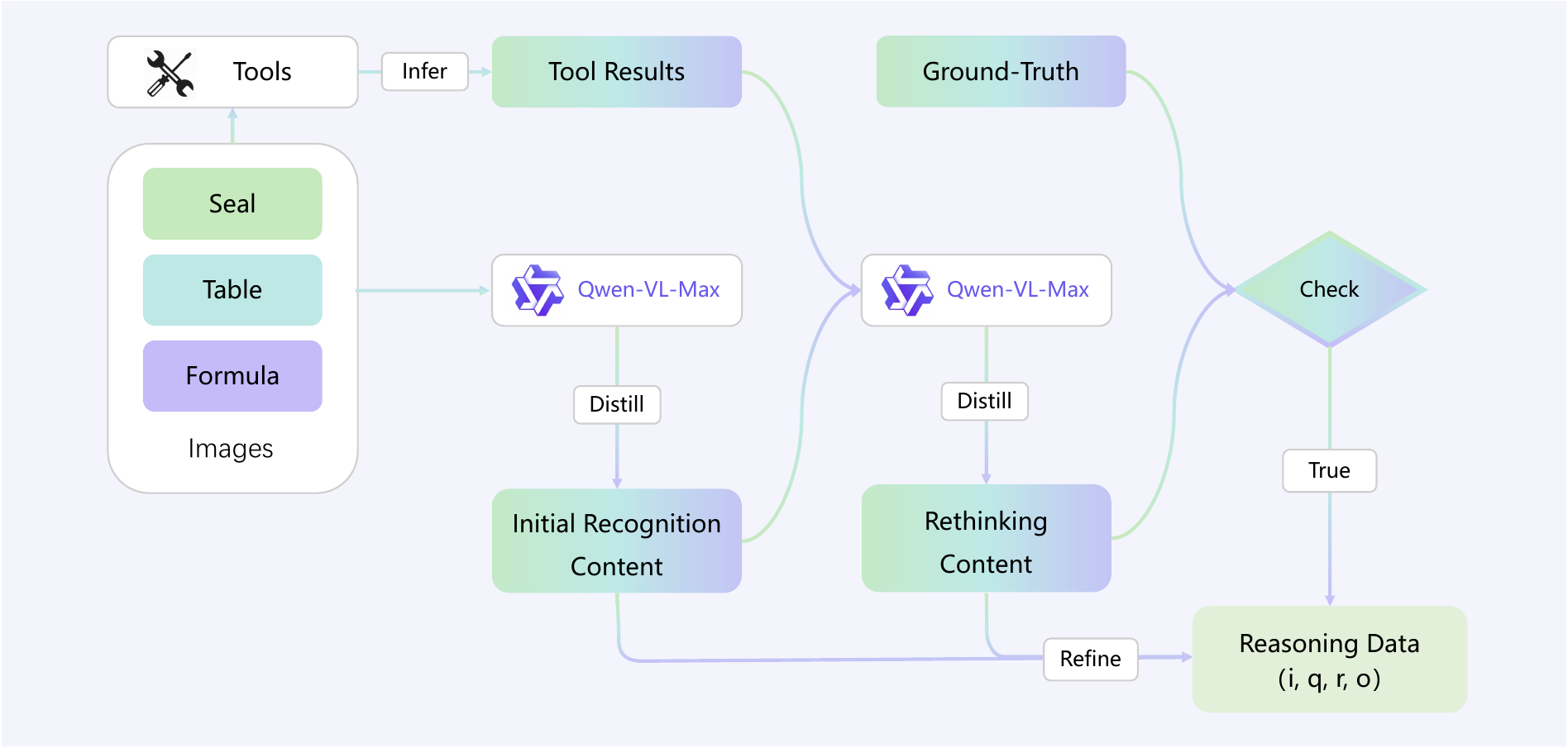}
    \caption{Overall pipeline for constructing reasoning data.}
    \label{fig:data}
\end{figure}

\subsection{Data Construction}
To prove the effectiveness of our method, we conduct experiments on three recognition tasks, i.e., seal, table, and formula recognition tasks. We denote our original seal, table, and formula recognition datasets as $D_{sr}/D_{tr}/D_{fr} = (i_i, q_i, a_i)$, where $i_i$ denotes the i-th seal, table, or formula image, $q_i$ refers to the instruction prompting the model to recognize the content, and $a_i$ corresponds to the seal, table (in HTML format), or formula (in LaTeX format) content. 

\Cref{fig:data} shows the pipeline for constructing reasoning data. Specifically, we leverage the advanced capabilities of powerful VLMs, such as Qwen-VL-Max\footnote{\url{https://help.aliyun.com/zh/model-studio/vision}}, through a well-designed prompt and an example to generate an interleaved reasoning-and-tool chain $r_i$ and an output $o_i$. The reasoning chain $r_i$ consists of three parts, content recognized by model itself enclosed within <think> and </think> tags, multiple tool responses enclosed within <tool> and </tool> tags, reflects enclosed within <rethink> and </rethink> tags. The generated output is enclosed within <answer> and </answer> tags. If the output $o_i$ matches the ground-truth $a_i$, we retain the instance $(i_i, q_i, r_i, o_i)$ as a valid reasoning sample, with which we ultimately construct three reasoning datasets $R_{sr}/R_{tr}/R_{fr} = (i_i, q_i, r_i, o_i)$. For example, as shown in \Cref{fig:prompt_1}, \Cref{fig:prompt_2} and \Cref{fig:prompt_3}, we use a multi-turn conversation approach to generate data. Specifically, in the first turn, we prompt the VLM to recognize the content in the input image. In the second turn, we provide results from other tools as references, allowing the VLM to analyze and give the final answer.

Considering the differences between three tasks, we select data from different sources and use different tools when constructing reasoning chains.
\begin{itemize}
    \item \textbf{Seal Recognition.} We select data from the ReST \cite{yu2023icdar} dataset. Since the annotations of the testing set are not publicly available, we only use the training set to verify our method. We find that there is considerable repetition in the seal titles. To prevent the language model part from memorizing fixed phrases in the seal titles, we isolate the data during splitting training and testing sets (using community detection for coarse filtering and specific keywords for fine filtering). When generating the reasoning chains, we use the results of PP-StructureV3 and Qwen-VL-OCR\footnote{\url{https://help.aliyun.com/zh/model-studio/qwen-vl-ocr}} as references. If the output $o$ and the answer $a$ match exactly in terms of characters, we retain the instance as a valid reasoning sample.
    
    \item \textbf{Table Recognition.} We select data from in-house dataset and the TabRecSet \cite{yang2023large} dataset. When generating the reasoning chain, we use the results of PP-StructureV3 and MonkeyOCR-3B\cite{li2025monkeyocr} as references. If the Tree-Edit-Distance-based Similarity (TEDS) \cite{zhong2020image} score between the output $o$ and the answer $a$ lager than the content recognized by model itself and exceeds a threshold (set to 0.98), we retain the instance as a valid reasoning sample.
    
    \item \textbf{Formula Recognition.} We select data from UniMER-1M \cite{wang2024unimernet} dataset. Similarly, we use the results of PP-StructureV3 and MonkeyOCR-3B as references in the reasoning chain. If the Normalized Edit Distance (NED) \cite{zhang2019icdar} score between the output $o$ and the answer $a$ smaller than the content recognized by model itself and is below a threshold (set to 0.015), we retain the instance as a valid reasoning sample.
\end{itemize}

\subsection{Model Training}
\subsubsection{Learning reasoning with SFT.} 
The reasoning datasets $R_{sr}$, $R_{tr}$, and $R_{fr}$ are utilized to fine-tune the base model, Qwen2.5-VL-Instruct, to generate a CoT followed by a final answer. Each training instance $(i, q, r, o)$ consists of an image $i$, an instruction $q$, a reasoning chain $r$ and a output $o$. During the SFT stage, the image $i$ and the instruction $q$ serve as the input to the model, while the reasoning chain $r$ and the output $o$ are treated as the target output, enabling the model to learn to produce coherent reasoning steps along with the correct solution.

\subsubsection{Learning reasoning with RFT.} We reinforcement fine-tune the base model using the same instances from $R_{sr}$, $R_{tr}$, and $R_{fr}$, except the reasoning chain $r$ and the output $o$. Specifically, during the RFT stage, the image $i$ and the instruction $q$ serve as the input to the model, while the answer $a$ are treated as the target output. In addition, we use a system prompt to guide the model to generate <think>, <rethink>, and <answer> content and provide <tool> content to it. We adopt the GRPO algorithm, incorporating two reward mechanisms: a format reward to ensure the generated output adheres to the desired structure, and an accuracy reward to encourage correct answers. 
\begin{itemize}
    \item \textbf{Format reward.} A reward score of 1.0 is granted if the CoT contains exactly <think> </think>, <tool> </tool>, <rethink> </rethink> and <answer> </answer> tags, with no additional content outside these boundaries. If this strict formatting criterion is not met, it receives a reward of 0.0.
    \item \textbf{Accuracy reward.} For seal recognition, if the output $o$ exactly matches the answer $a$, the model receives a reward score of 1.0; otherwise, it receives a score of 0.0. For table recognition, the TEDS score between $o$ and $a$ is used as the reward score. For formula recognition, the Character Detection Matching (CDM) \cite{wang2024cdm} metric between $o$ and $a$ serves as the reward score, and an additional 0.5 will be added if the score is equal to 1.0.
\end{itemize}

\section{Experiment}
\label{sec:experiment}

\begin{table}[t]
    \caption{Statistics of the training sets.}
    \label{tab:train}
    \centering
    \begin{tabularx}{\textwidth}{l|*{1}{>{\centering\arraybackslash}X}|*{1}{>{\centering\arraybackslash}X}|*{1}{>{\centering\arraybackslash}X}|*{1}{>{\centering\arraybackslash}X}|*{1}{>{\centering\arraybackslash}X}}
        \toprule
        \bf Datasets & \bf Size & \bf $q_{token}$ & \bf $i_{token}$ & \bf $r_{token}$ & \bf $a_{token}$ \\
        \midrule
        $R_{sr}$ & 1024 & 54 & 106 & 417 & 14 \\
        $R_{tr}$ & 1024 & 15 & 369 & 1697 & 324 \\
        $R_{fr}$ & 1024 & 17 &54 & 769 & 65 \\
        \bottomrule
    \end{tabularx}
\end{table}

\begin{table}[t]
    \caption{Statistics of the testing sets.}
    \label{tab:test}
    \centering
    \begin{tabularx}{\textwidth}{l|*{1}{>{\centering\arraybackslash}X}|*{1}{>{\centering\arraybackslash}X}}
    \toprule
    \bf Dataset & \bf Language & \bf Size \\
    \midrule
    ReST* & Chinese & 432 \\
    OmniDocBench(Table) & Chinese/English/Mixed & 428 \\
    OmniDocBench(Formula) & Chinese/English & 353 \\
    \bottomrule
    \end{tabularx}
\end{table}

\subsection{Experimental Setups}
\subsubsection{Training.} \Cref{tab:train} shows the statistics of the training sets. We train DianJin-OCR-R1 models on a single node with 8 NVIDIA A100 GPUs. In the SFT stage, we train models over 2 epochs, using a learning rate of $1.0\times10^{-5}$, 1 training batch size per device and gradient accumulation with 4 steps. In the RFT stage, we perform 8 rollouts per sample, with a train batch size of 128 and a rollout batch size of 32. We use a learning rate of $1.0\times10^{-6}$ and a sampling temperature of 1.0 over 3 epochs.

\subsubsection{Evaluation.} We evaluate our models on multiple benchmarks, including the ReST testing sets, the OmniDocBench table sets, and the OmniDocBench formula sets. They include multiple languages, such as Chinese, English, and a mix of both. The detailed statistics of these testing sets are summarized in \Cref{tab:test}. For the ReST testing sets, the results are evaluated by accuracy. For the OmniDocBench table sets, we report the TEDS, structure TEDS (STEDS), and Normalized Edit Distance. For the OmniDocBench formula sets, we evaluate the results using CDM, ExpRate@CDM and Normalized Edit Distance. 

\subsubsection{Baselines.} We compare our model against three categories of models. The first category includes General VLMs: Qwen2.5-VL-7B-Instruct\cite{bai2025qwen2}, and Intern3VL-8B\cite{zhu2025internvl3}. The second category consists of Expert VLMs, including Qwen-VL-OCR, GOT\cite{wei2024general}, and MonkeyOCR-3B\cite{li2025monkeyocr}. The third category includes Expert OCR Models: PP-StructureV3\footnote{PP-OCRv4\_server\_seal\_det + PP-OCRv4\_server\_rec} for seal recognition, PP-StructureV3\footnote{PP-LCNet\_x1\_0\_table\_cls + SLANeXt\_wired + SLANeXt\_wireless} and RapidTable\footnote{SLANet\_plus, \url{https://github.com/RapidAI/RapidTable}} for table recognition, and PP-StructureV3\footnote{PP-FormulaNet\_plus-M} and UniMERNet-B\cite{wang2024unimernet} for formula recognition.

\subsection{Experimental Results}
\begin{table}[t]
    \caption{Performance comparison across different testing sets. Scores in \textbf{bold} and \underline{underlined} indicate the best and second-best results, respectively.}
    \label{tab:results}
    \centering
    \begin{tabularx}{\textwidth}{l|*{1}{>{\centering\arraybackslash}X}|*{3}{>{\centering\arraybackslash}X}|*{3}{>{\centering\arraybackslash}X}}
        \toprule
        \multirow{2}{*}{\bf Model} & \bf Seal & \multicolumn{3}{c|}{\bf Table} & \multicolumn{3}{c}{\bf Formula} \\
        \cmidrule{2-8}
        & ACC$\uparrow$ & TEDS$\uparrow$ & STEDS$\uparrow$ & NED$\downarrow$ & CDM$\uparrow$ & ER$\uparrow$ & NED$\downarrow$ \\
        \midrule
        \rowcolor[gray]{0.8}
        \multicolumn{8}{c}{Expert OCR Models} \\
        \midrule
        PP-StructureV3 & 0.600 & 0.818 & 0.918 & 0.121 & 0.966 & 0.730 & 0.198  \\
        RapidTable & - & 0.832 & 0.902 & 0.120 & - & - & - \\
        UniMERNet-B & - & - & - & - & 0.932 & 0.645 & 0.290  \\
        \midrule
        \rowcolor[gray]{0.8}
        \multicolumn{8}{c}{Expert VLMs} \\
        \midrule
        GOT & - & 0.787 & 0.885 & 0.219 & 0.898 & 0.622 & 0.287 \\
        MonkeyOCR-3B & - & 0.845 & 0.895 & 0.182 & 0.973 & 0.767 & 0.196 \\
        Qwen-VL-OCR & 0.472 & - & - & - & - & - & -\\
        \midrule
        \rowcolor[gray]{0.8}
        \multicolumn{8}{c}{General VLMs} \\
        \midrule
        Qwen2.5-VL-7B-Instruct & 0.527 & 0.853 & 0.904 & 0.107 & 0.915 & 0.642 & 0.243 \\
        InternVL3-8B & 0.475 & 0.812 & 0.880 & 0.139 & 0.946 & 0.668 & 0.273 \\
        \midrule
        \rowcolor[gray]{0.8}
        \multicolumn{8}{c}{Dianjin-OCR-R1} \\
        \midrule
        Dianjin-OCR-R1(SFT) & \underline{0.722} & \underline{0.895} & \underline{0.935} & \underline{0.080} & \textbf{0.977} & \underline{0.770} & \underline{0.188} \\
        Dianjin-OCR-R1(RFT) & \textbf{0.766} & \textbf{0.901} & \textbf{0.939} & \textbf{0.072} & \underline{0.976} & \textbf{0.787} & \textbf{0.179} \\
        \bottomrule
    \end{tabularx}
\end{table}

\Cref{tab:results} compares the performance of our DinJin-OCR-R1 models with other models. The results reveal the following key points:
\begin{itemize}
    \item Our DianJin-OCR-R1 models significantly outperform the base model on the three testing sets. For instance, models trained by SFT improves accuracy from 0.527 to 0.722 on seal recognition task, TEDS from 0.853 to 0.895 on table recognition task, and CDM from 0.915 to 0.977 on formula recognition task. These results indicate that the proposed method can effectively improve OCR performance.
    \item Our DianJin-OCR-R1 models outperform the expert OCR models used in the reasoning chains, which indicates that our models not only effectively leverage the results of the expert OCR models, but also surpasses them. 
    \item Our DianJin-OCR-R1 models achieve best results with the model using only thousands of training samples. Using the same samples, the models trained by RFT are generally superior to those trained by SFT, indicating that RFT is more advantageous in our experiments.
\end{itemize}

\begin{table}[t]
    \centering
    \caption{Performance comparison with and without tools.}
    \label{tab:ablation}
    \begin{tabularx}{\textwidth}{l|*{1}{>{\centering\arraybackslash}X}|*{3}{>{\centering\arraybackslash}X}|*{3}{>{\centering\arraybackslash}X}}
        \toprule
        \multirow{2}{*}{\bf Model} & \bf Seal & \multicolumn{3}{c|}{\bf Table} & \multicolumn{3}{c}{\bf Formula} \\
        \cmidrule{2-8}
        & ACC$\uparrow$ & TEDS$\uparrow$ & STEDS$\uparrow$ & NED$\downarrow$ & CDM$\uparrow$ & ER$\uparrow$ & NED$\downarrow$ \\
        \midrule
        Qwen2.5-VL-7B-Instruct & 0.527 & 0.853 & 0.904 & 0.107 & 0.915 & 0.642 & 0.243 \\
         + tools & \textbf{0.560} & \textbf{0.865} & \textbf{0.908} & \textbf{0.107} & \textbf{0.958} & \textbf{0.713} & \textbf{0.241} \\
        \midrule
        SFT & 0.643 & 0.863 & 0.910 & 0.098 & 0.938 & 0.599 & 0.196 \\
        DianJin-OCR-R1(SFT) & \textbf{0.722} & \textbf{0.895} & \textbf{0.935} & \textbf{0.080} & \textbf{0.977} & \textbf{0.770} & \textbf{0.188} \\
        \midrule
        RFT & 0.634 & 0.897 & 0.937 & 0.078 & 0.974 & 0.727 & 0.204\\
        DianJin-OCR-R1(RFT) & \textbf{0.766} & \textbf{0.901} & \textbf{0.939} & \textbf{0.072} & \textbf{0.976} & \textbf{0.787} & \textbf{0.179} \\
        \bottomrule
    \end{tabularx}
\end{table}

\begin{table}[t]
    \centering
    \caption{Performance comparison of models using different tools.}
    \label{tab:tools}
    \begin{tabularx}{\textwidth}{l|*{1}{>{\centering\arraybackslash}X}|*{1}{>{\centering\arraybackslash}X}|*{1}{>{\centering\arraybackslash}X}}
        \toprule
        \bf Model & \bf CDM$\uparrow$ & \bf ER$\uparrow$ & \bf NED$\downarrow$ \\
        \midrule
        Tool1: GOT & 0.898 & 0.622 & 0.287 \\
        Tool2: PP-FormulaNet\_plus-S & 0.852 & 0.523 & 0.299 \\
        DianJin-OCR-R1(SFT) & \textbf{0.969} & \textbf{0.722} & \textbf{0.227}  \\
        \midrule
        \midrule
        Tool1: MonkeyOCR-3B & 0.973 & 0.767 & 0.196 \\
        Tool2: PP-FormulaNet\_plus-M & 0.966 & 0.730 & 0.198 \\
        DianJin-OCR-R1(SFT) & \textbf{0.977} & \textbf{0.770} & \textbf{0.188} \\
        \bottomrule
    \end{tabularx}
\end{table}

\subsection{Analysis}
\subsubsection{Effectiveness of Tool Use.} We conduct three groups of comparative experiments: (1) providing tool results to the base model, Qwen2.5-VL-7B-Instruct, and prompt it to generate the final results via a two-turn conversation (same with data construction); (2) training the base model by SFT directly without using any tools; and (3)  training the base model by RFT directly without using any tools. \Cref{tab:ablation} summarizes our findings:
\begin{itemize}
    \item Providing tool results as references to the base model significantly boosts its performance, which demonstrates that leveraging other tools can effectively enhance the capabilities of the VLM on OCR tasks.
    \item Comparing to models trained by SFT and RFT directly without using any tools, our models incorporating reasoning chains and tool use consistently achieve superior performance, which highlights the successful integration of reasoning and tools.
\end{itemize}

\subsubsection{Impact of Using Different Tools.} We also evaluate how using different tools influences the overall performance. Taking the formula recognition task as an example, we additionally employ GOT and PP-FormulaNet\_plus-S as tools. As shown in \Cref{tab:tools}, the performance using MonkeyOCR-3B and PP-FormulaNet\_plus-M are better than using GOT and PP-FormulaNet\_plus-S, which indicates that leveraging stronger expert models or iterating expert models can improve overall performance. Compared with fine-tuning VLMs, this approach entails substantially lower cost.

\begin{figure}[t]
  \centering
  \begin{subfigure}[t]{0.45\textwidth}
    \centering
    \includegraphics[width=\linewidth]{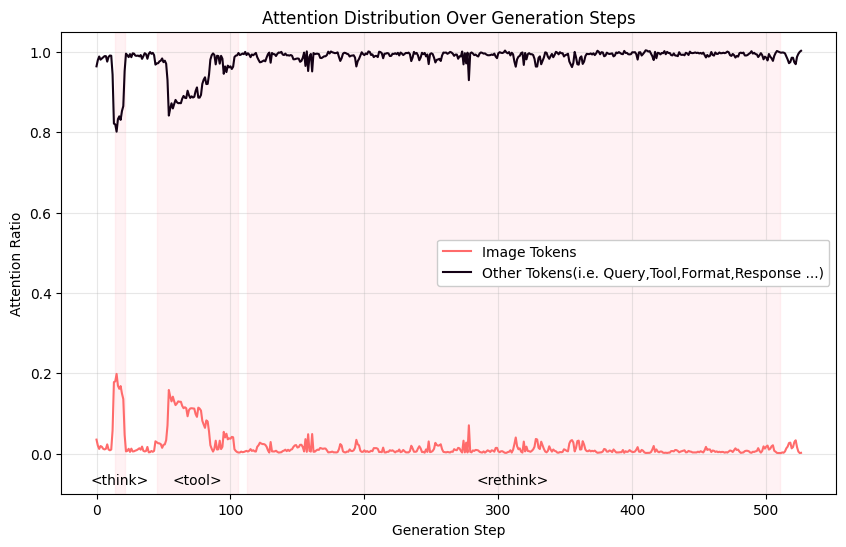}
    \caption{DianJin-OCR-R1(SFT)}
    \label{fig:b}
  \end{subfigure}\hfill
  \begin{subfigure}[t]{0.45\textwidth}
    \centering
    \includegraphics[width=\linewidth]{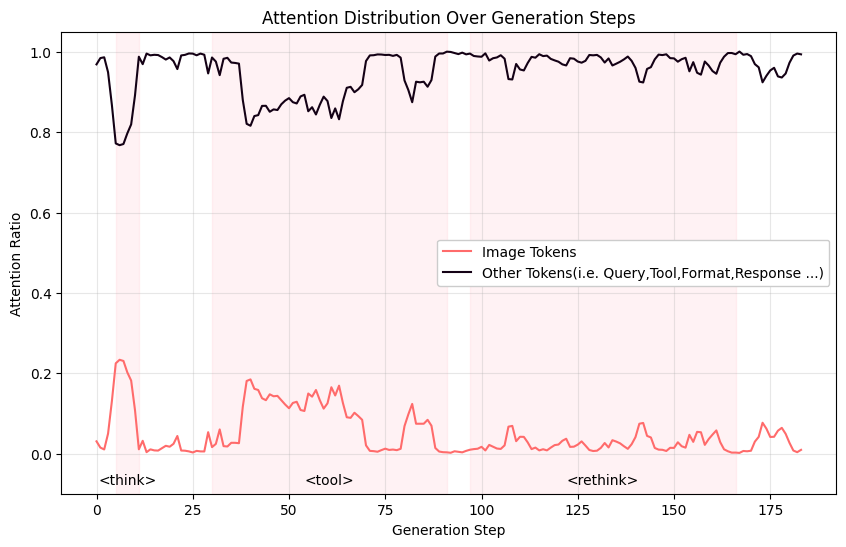}
    \caption{DianJin-OCR-R1(RFT)}
    \label{fig:c}
  \end{subfigure}
  \caption{Visualization of look-again. The red curve denotes the attention between each token and the image tokens, whereas the black curve denotes the attention between each token and the other tokens.}
  \label{fig:look}
\end{figure}

\begin{table}[t]
\caption{Look-again radio.}
\label{tab:radio}
\centering
\begin{tabular}{l|c}
\toprule
\textbf{Models} & \textbf{Radio} \\
\midrule
DianJin-OCR-R1(SFT) & 81.02\% \\
DianJin-OCR-R1(RFT) & 90.97\%  \\
\bottomrule
\end{tabular}
\end{table}

\subsubsection{Look-again.}
To verify whether our models genuinely "look again" at the image during the <rethink> stage, we conduct further experiments. First, we compute the attention between each token and the image tokens. As shown in \Cref{fig:look}, our models successfully and repeatedly re-focus on the visual input (spikes in the red curve) during the <rethink> stage, with this phenomenon being more pronounced in the DianJin-OCR-R1(RFT). In addition, we further define the look-again radio as the proportion of samples whose attention curves exhibit spikes relative to the total number of samples (using find\_peaks function in the scipy package to find spikes), and evaluate it on the ReST testing sets. As shown in \Cref{tab:radio}, the radio of DianJin-OCR-R1(RFT) is higher than DianJin-OCR-R1(SFT), indicating that DianJin-OCR-R1(RFT) re-focuses more on the visual input. All the above observations are consistent with the conclusion that DianJin-OCR-R1(RFT) achieves better performance than DianJin-OCR-R1(SFT).

\section{Conclusion}
\label{sec:conclusion}
We have presented DianJin-OCR-R1, a reasoning-augmented framework for large vision-language models on OCR tasks. This framework fosters a straightforward reasoning paradigm in VLMs, enabling a more transparent and verifiable reasoning process on complex document OCR tasks. Through extensive experiments on diverse benchmarks, we demonstrate that reasoning-aware training significantly improves performance.

\section*{Acknowledgements}
We extend our sincere thanks to other members of the Qwen DianJin Team for their significant contributions, whose hard work and dedication were crucial to the success of this project.

\bibliography{colm2024_conference}
\bibliographystyle{colm2024_conference}

\appendix
\section{Prompts used in this paper}
\label{apdx:prompts}

\begin{figure*}[h]
\begin{tcolorbox}[title=Prompt Template, colback=gray!5, colframe=black, fonttitle=\bfseries]
\begin{verbatim}
## Turn 1
你是一名专业的OCR专家，你的任务是识别图片中的印章抬头。印章抬头位于印章上方，通常用于标明印章的归属、用途或机构名称，是印章的重要组成部分，帮助识别印章的合法性和适用范围。请将你的识别结果放在<recognition></recognition>之间。

## Turn 2
以下是其它工具对该印章的识别内容：
<tool>
{
    "ocr_tool_1": "{tool1}",
    "ocr_tool_2": "{tool2}" 
}
</tool>
请比较你初步识别的结果和其它工具识别内容之间的**全部**不同之处。当存在差异时，你需要重新查看图片，并分析你的初步识别结果是否错误或者是否有遗漏。（你跟其他 OCR 工具均有可能犯错。请务必再次仔细查看图片）
将你的对比与分析过程放在<rethink></rethink>之间，并将最终识别的**印章抬头**内容放在<answer></answer>之间。
\end{verbatim}

\end{tcolorbox}
\caption{Prompt used to construct reasoning data for seal recognition.}

\label{fig:prompt_1}
\end{figure*}

\begin{figure*}[htbp]
\begin{tcolorbox}[title=Prompt Template, colback=gray!5, colframe=black, fonttitle=\bfseries]
\begin{verbatim}
## Turn 1
请将图中的表格以 HTML 格式输出。

## Turn 2
以下是其它工具识别的 HTML 格式的内容：
<tool>
## 工具1：
{tool1}
## 工具2：
{tool2}
</tool>
请重新仔细查看图片，比较之前识别的结果和其它工具结果，分析表格结构(如单元格是否和图中一致、合并单元格 colspan 和 rowspan 是否正确等)和表格内容(如单元格中的内容是否识别错误或有遗漏)。如果有合并单元格，请以 colspan="x"/rowspan="x" 表示。
将分析过程放在<rethink></rethink>之间，并将最终识别的 HTML 放在<answer></answer>之间。

## 输出格式：
<rethink>
...
</rethink>
<answer>
HTML
</answer>
\end{verbatim}

\end{tcolorbox}
\caption{Prompt used to construct reasoning data for table recognition.}
\label{fig:prompt_2}
\end{figure*}

\begin{figure*}[htbp]
\begin{tcolorbox}[title=Prompt Template, colback=gray!5, colframe=black, fonttitle=\bfseries]
\begin{verbatim}
## Turn 1
请用 LaTeX 格式写出图中公式的表达式。

## Turn 2
以下是其它工具识别的 LaTeX 格式的内容：
<tool>
## 工具1结果：
{tool1}
## 工具2结果：
{tool2}
</tool>
请重新仔细查看图片，比较之前识别的结果和其它工具结果之间的不同之处，分析是否识别错误或有遗漏。
将分析过程放在 <rethink></rethink> 之间，并将最终识别的 LaTeX 放在 <answer></answer> 之间。

## 输出格式：
<rethink>
...
</rethink>
<answer>
LaTeX
</answer>
\end{verbatim}

\end{tcolorbox}
\caption{Prompt used to construct reasoning data for formula recognition.}
\label{fig:prompt_3}
\end{figure*}

\end{CJK*}
\end{document}